\documentclass[]{spie}  %>>> use for US letter paper
\usepackage{amsmath,amsfonts,amssymb}
\usepackage{graphicx}
\usepackage[colorlinks=true, allcolors=blue]{hyperref}
\usepackage{lineno,hyperref}

\usepackage{float}
\usepackage{lineno,hyperref}
\usepackage{array,color}
\usepackage{algorithm}
\usepackage{algorithmic}
\usepackage{epstopdf}
\usepackage{subfigure}
\usepackage{multicol}

\title{Volume Calculation of CT lung Lesions based on Halton Low-discrepancy Sequences}

\author[a]{Liansheng Wang}
\author[a]{Shusheng Li}
\author[b]{Shuo Li}
\affil[a]{Department of Computer Science, Xiamen University, Xiamen, China}
\affil[b]{Dept. of Medical Imaging, Schulich School of Medicine and Dentistry, University of Western Ontario, Canada}

\authorinfo{Further author information: (Send correspondence to Liansheng Wang)\\Liansheng Wang: E-mail: lswang@xmu.edu.cn}

\begin{document}
\maketitle

\begin{abstract}
Volume calculation from the Computed Tomography (CT) lung lesions data is a significant parameter for clinical diagnosis. The volume is widely used to assess the severity of the lung nodules and track its progression, however, the accuracy and efficiency of previous studies are not well achieved for clinical uses. It remains to be a challenging task due to its tight attachment to the lung wall, inhomogeneous background noises and large variations in sizes and shape.

In this paper, we employ Halton low-discrepancy sequences to calculate the volume of the lung lesions. The proposed method directly compute the volume without the procedure of three-dimension (3D) model reconstruction and surface triangulation, which significantly improves the efficiency and reduces the complexity. The main steps of the proposed method are: (1) generate a certain number of random points in each slice using Halton low-discrepancy sequences and calculate the lesion area of each slice through the proportion; (2) obtain the volume by integrating the areas in the sagittal direction. In order to evaluate our proposed method, the experiments were conducted on the sufficient data sets with different size of lung lesions. With the uniform distribution of random points, our proposed method achieves more accurate results compared with other methods, which demonstrates the robustness and accuracy for the volume calculation of CT lung lesions. In addition, our proposed method is easy to follow and can be extensively applied to other applications, e.g., volume calculation of liver tumor, atrial wall aneurysm, etc.
\end{abstract}


\section{Purpose}
\label{sec:intro}  % \label{} allows reference to this section

Volume calculation of lung lesions from Computed Tomography (CT) images plays a vital role in the evaluation of the level of lung disease. An efficient and accuracy method for volume measurement can provide reliable information for clinical diagnosis, which helps doctors to evaluate the state of patients and make appropriate therapies. Furthermore, doctors can estimate the timeliness and treatment effect of drugs with the help of the volume of lung lesions~\cite{Wang2014Direct}.

However, many previous studies about the volume calculation for lesions tissue or organ from CT or MRI images are based on the diameters of lung lesions which roughly expresses the size, or based on 3D model reconstruction and surface triangulation which are not efficient and sometimes fail to build a perfect 3D model due to the high complexity of medical images~\cite{Punithakumar2010Detection}.

Monte Carlo method (MC) is very useful tool for physical and mathematical problems, especially in optimization, numerical integration~\cite{Afshin2014Regional}. Quasi-Monte Carlo methods (qMC) (also called low-discrepancy sequences), which were proposed to generate random points with better uniformity and lower discrepancy, are often considered as a replacement of MC. Some qMC methods have already been used for computing the volume~\cite{Zhen2014Direct}.

In this paper, to accurately and directly calculate the volume of lung lesions, we introduce the low-discrepancy sequences to compute volume. In our study, the Halton low-discrepancy sequences are generated to calculate the area of each CT slice and the volume are obtained by accumulating these slices' area with a smooth formula. Experiments are conducted to evaluate our method compared with other methods.

\section{Method}

\begin{figure}[H]
\centering
\subfigure[pseudorandom number sequence]{\includegraphics[height=1.7in,width=2.2in]{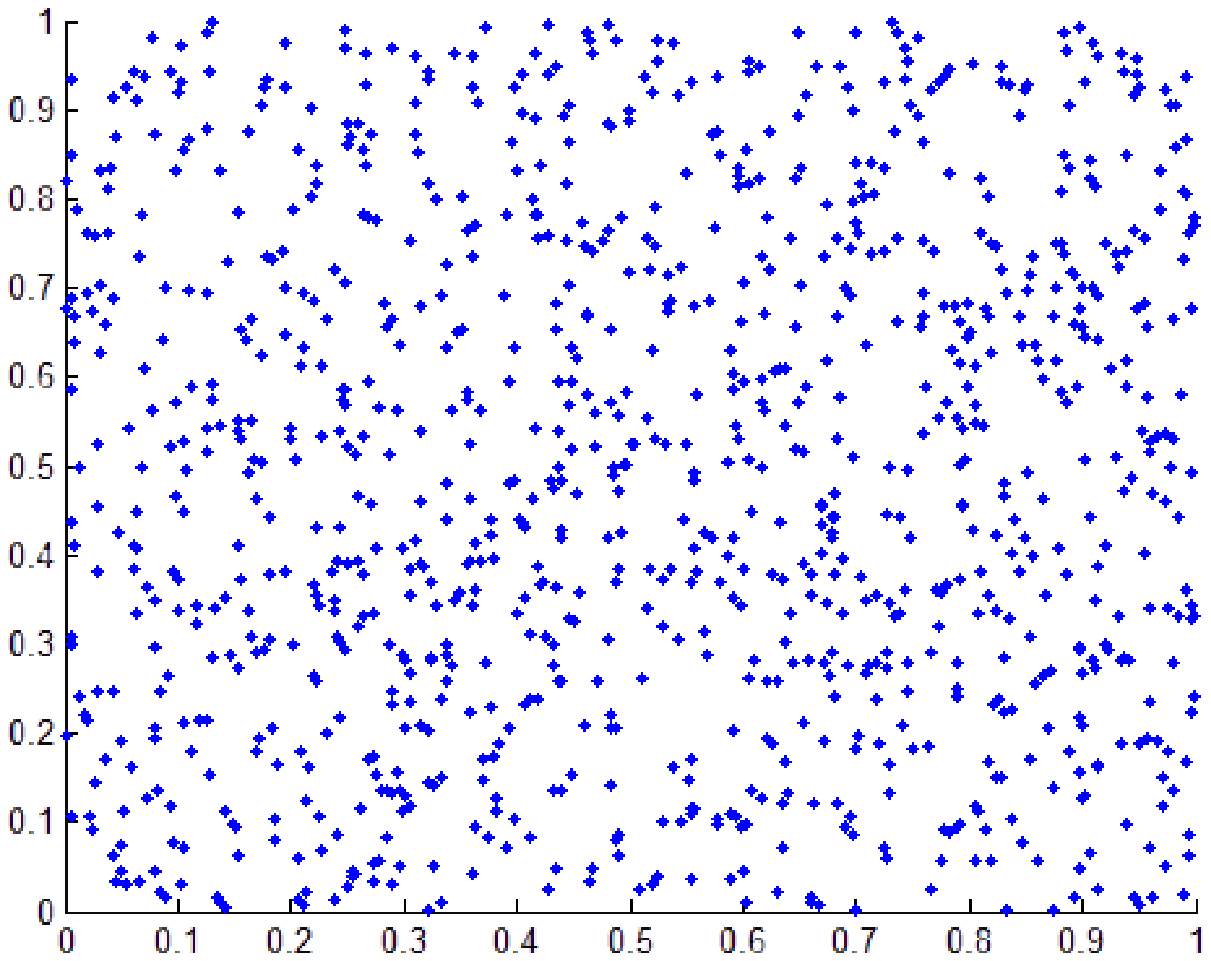}}\hspace{0.7cm}
\subfigure[Halton sequence]{\includegraphics[height=1.7in,width=2.2in]{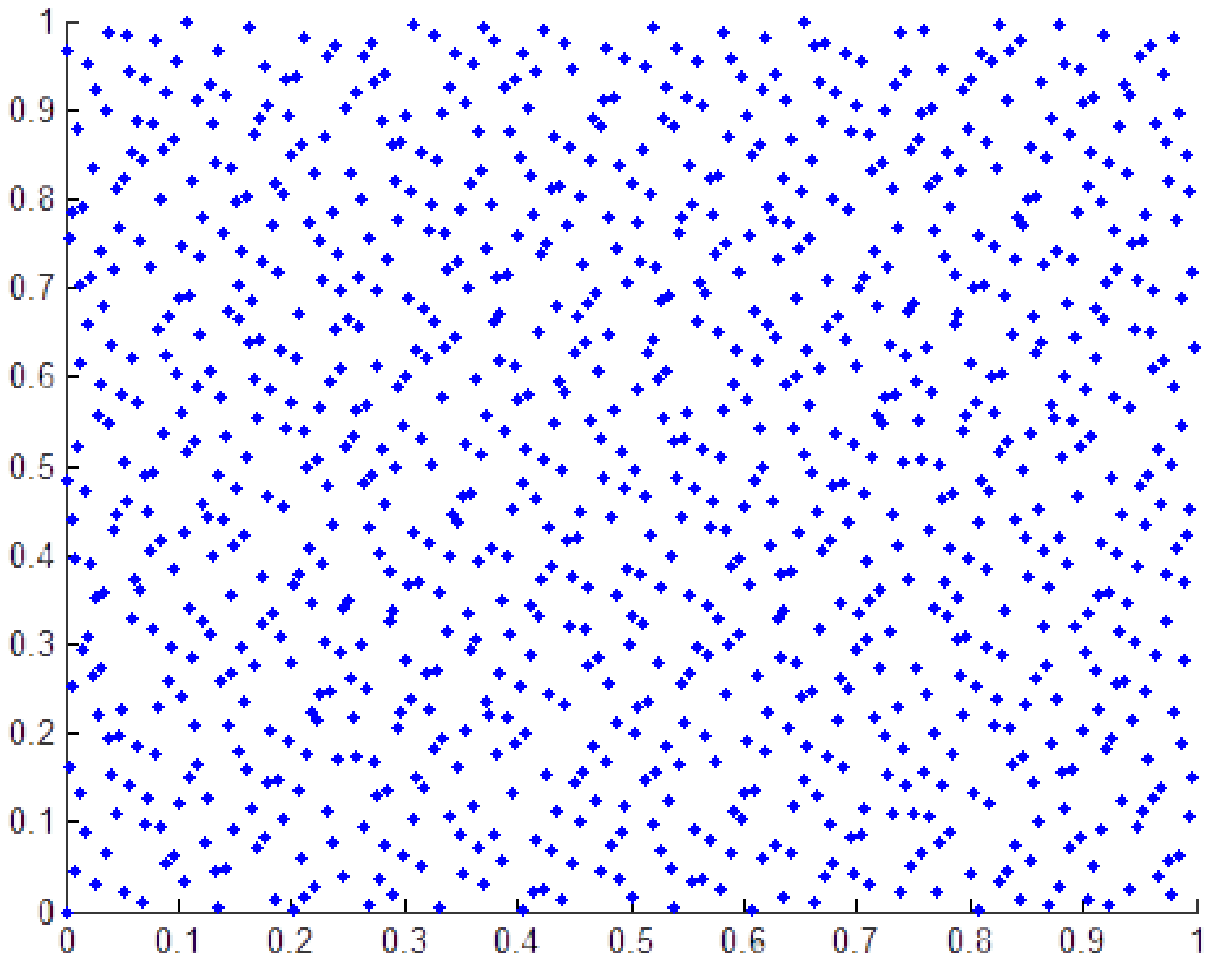}}
\caption{2-D plots of 1000 points generated respectively by (a) the pseudo-random number sequence, and (b) the Halton sequence.}\label{fig:compare}
\end{figure}

MC method is a simple and efficient numerical method based on random sampling. However, in \cite{davies1998low}, Davies \emph{et al.} pointed out that MC method is subject to errors and in particular all parts of space can not guarantee the equal sampling. Therefore, the sequence of Monte Carlo is not entirely uniform and it may lead to inaccuracy results. Fig  \ref{fig:compare}(a) illustrates the distribution of pseudorandom number tends to be lumped.

To generate more uniform sequences, low-discrepancy sequences are proposed. Compared with MC method, quasi-Monte Carlo method solves numerical problems using low-discrepancy sequences. The distribution of the Halton sequence is more uniformly as shown in Fig \ref{fig:compare}(b). Low-discrepancy sequences also have lower statistical errors. Low-discrepancy sequences have theoretical error bound O($N^{-1}$$(logN)^{D}$) compared with pseudorandom sequences O($N^{-1/2}$) \cite{davies1998low}, where $N$ is the number of samples and $D$ denotes the dimensions. In this paper, we employ Halton sequences to generate quasi-random numbers. The Halton sequences are constructed according to a deterministic method that uses a prime number as its base. The sequence is implemented easily and has a very uniform distribution on low dimension.

\begin{figure}[H]
  \centering
  \includegraphics[width=2in]{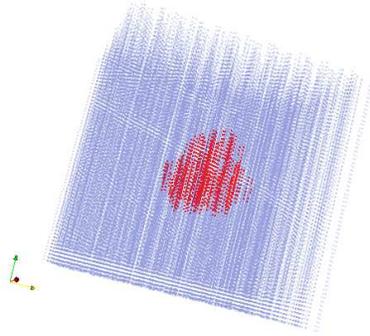}\\
  \caption{The distribution of Halton sequence in the 3D CT lung scans.}\label{fig:volumepoints}
\end{figure}

\textbf{Lesion Segmentation and Volume Calculation}. Our previous segmentation work \cite{wang20143d}  is introduced to obtain the segmented results. The procedure of volume calculation has five steps: (1) building reference coordinate system in every slice of CT scanning image; (2) generating a certain number of random points in each slice using Halton low-discrepancy sequences; (3) finding out the number of random points who are inside the lung lesion region; (4) calculating the area ($S_i$) of the lung lesion region for each slice by the proportion of random points to all random points, the area can be calculated by ${S_i} \approx \frac{m}{N}{S_N}$ where $m$ means the number of points inside the lung lesion, $N$ represents the number of all points generated, and $S_N$ is the area of each slice; and (5) calculating the volume of lung lesion using the formula of the frustum model as:
${\rm{V}} = \sum\limits_{i = 0}^{z - 1} {\frac{{({S_i} + {S_{i + 1}} + \sqrt {{S_i}{S_{i + 1}}} ) \times h}}{3}}$, where $S_i$ and $S_{i+1}$ define the areas of lesions in the $i$th slice and the $(i+1)$th slice, and $h$ denotes the slice thickness. Fig \ref{fig:volumepoints} illustrates the random points generated by Halton sequence in 3D in the CT lung scans.

\section{Experimental Results}
\label{sec:sections}
\begin{figure}[H]
  \centering
  \includegraphics[width=2in]{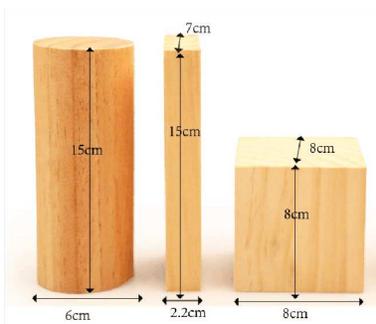}\\
  \caption{Illustration of regular models' parameters.}\label{fig:RegularModels}
\end{figure}

In this section, experiments are conducted on two kinds of CT data sets. The data was acquired at the 174 Hospital's CT room in Xiamen. The first data set is regular model, including cube, cuboid, and cylinder. See \ref{fig:RegularModels} for an illustration of the parameters of regular models, and we can calculate volumes of each model easily. Next, the CT scan data of each model was obtained using CT scanner in the CT room. And then the data can be used on our experiment. Table \ref{fig:regular} shows volumes of three regular models, which obviously demonstrates the result of three methods compared with ground truth volume. It has also shown that our method provides more accurate and precise results.

The second data set is manual tumor delineation of different size of lung lesions. Therefore, the data sets are divided into three groups: small, medium and large lung lesions. Each group has three data sets. The ground truth of each data set is provided by the experienced doctors who manually tag the CT lesion in each slice. The binary image of each slice is got using two value ($-4.0$ standing for background and $4.0$ indicating the lung lesion). This can guarantee the accurate segmentation of the lung lesion at the most extent. In order to evaluate our method, our results are compared with the Monte Carlo method and the method named LSTK using a discretized version of the divergence theorem described by Alyassin et al.\cite{alyassin1994evaluation}. Our experiments were run in Visual Studio 2010 using MicroSoft Windows 8.1 platform on a CPU of Intel 3.50GHz core i3-4150 with 4GB of RAM. In our proposed method, the dimension was set as $5$ to generate $5$ columns of low-discrepancy value and we chose the first and the fifth column (dimension) of the sequences.

We define the relative error $E$ as the evaluation criteria by the following formula:

\[E = \frac{{\left| {ComputedVolume - TrueVolume} \right|}}{{TrueVolume}}\]

\begin{table*}[ht]
\centering
\begin{tabular}{l|ccccccccc}
\hline
\hline
Dataset  &True Volume &LSTK (-0.5) &MC (10000) &Ours (10000)\\ \hline
Cube &512.000 &516.356(0.85\%) &510.732(0.25\%) &512.790(\textbf{0.15\%})\\
Cuboid &231.000 &232.898(0.82\%) &230.177(0.36\%) &230.788(\textbf{0.09\%})\\
Cylinder &424.115 &427.033(0.69\%) &421.189(0.69\%) &425.106(\textbf{0.23\%})\\
\hline
\end{tabular}
\caption{The results of three regular models' volume calculation of three methods compared to ground truth volume. The value of each volume is in $cm^3$. MC and our method use 10000 random points. LSTK uses the iso-surface threshold $-0.5$ at its best. The relative error of each volume calculation is shown in brackets after the volume value.}
\label{fig:regular}
\end{table*}

Table \ref{fig:table} shows the results of volume calculation and its corresponding relative errors for all methods. It is demonstrated that the results of LSTK have a lower accuracy and MC method's results is unstable. However, our method achieve more accurate results greatly closed to the true volume at most cases and more stable than MC method.

\begin{figure*}[ht]
\centering
\subfigure[cube]{\includegraphics[height=1in,width=2in]{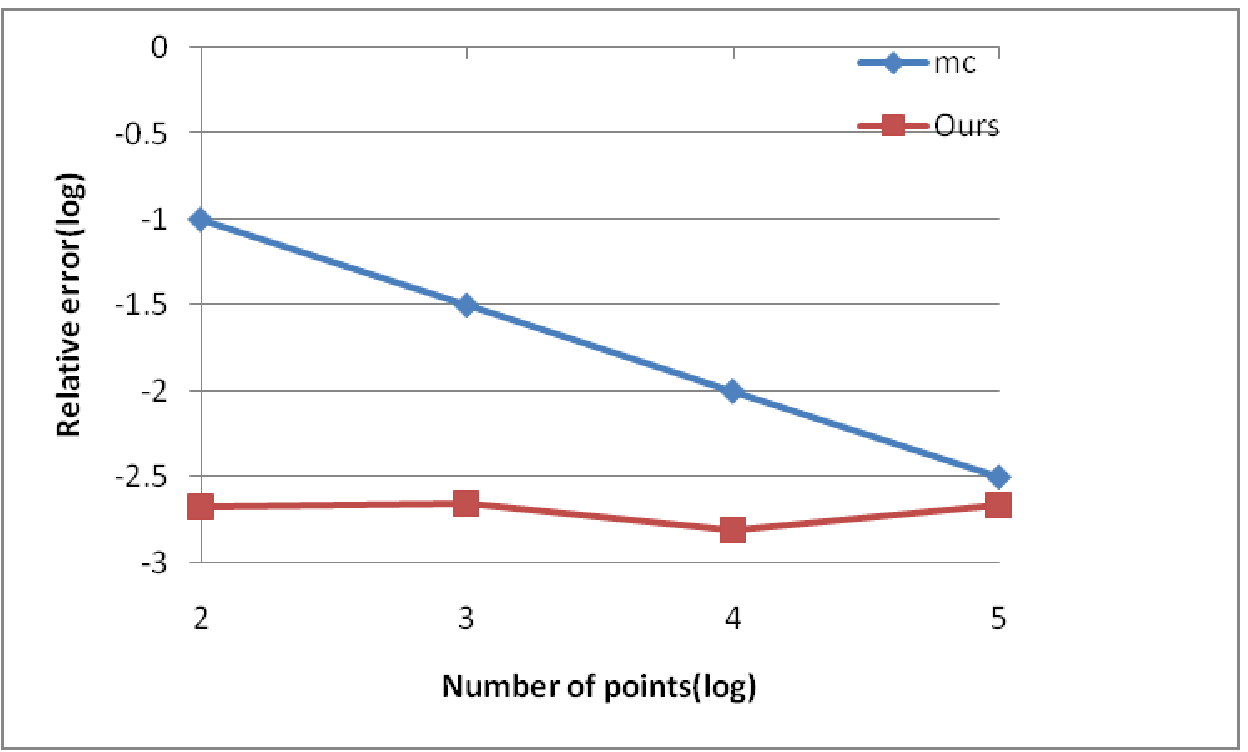}}
\subfigure[cuboid]{\includegraphics[height=1in,width=2in]{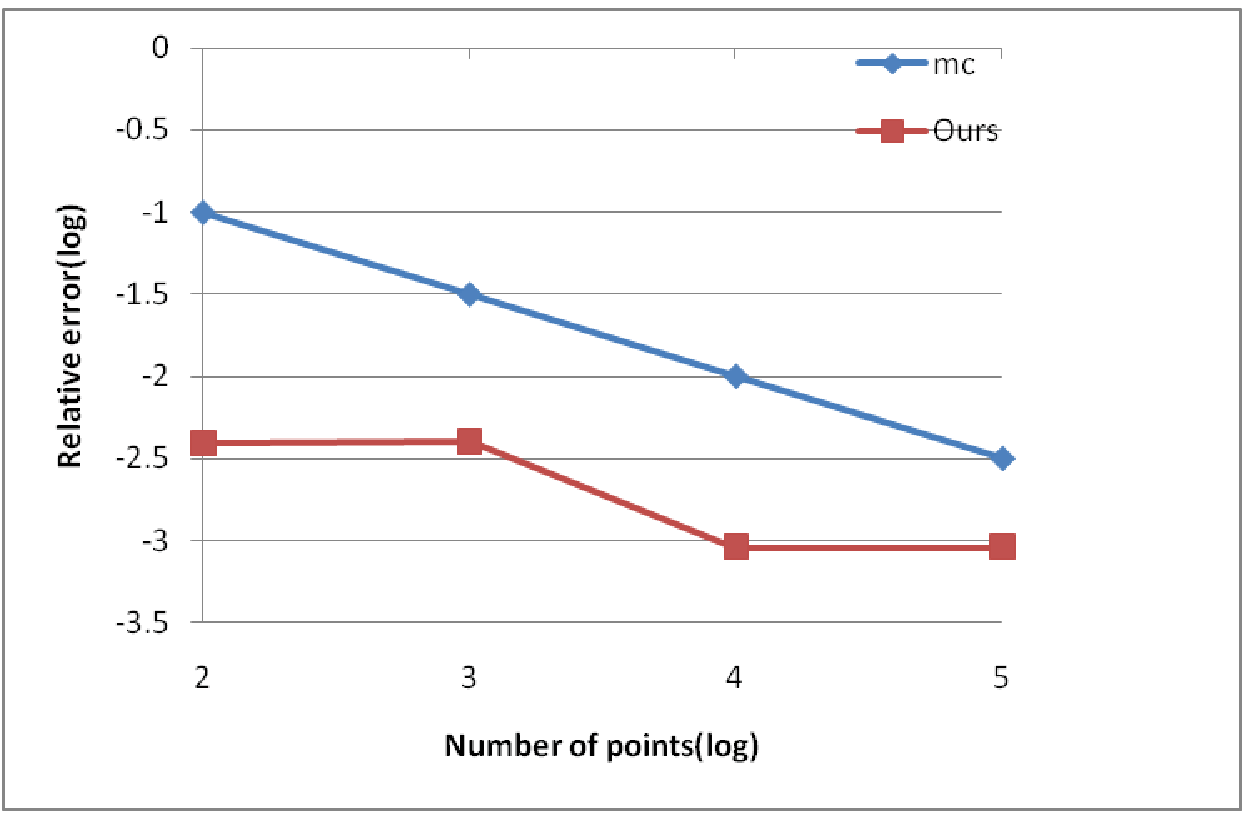}}
\subfigure[cylinder]{\includegraphics[height=1in,width=2in]{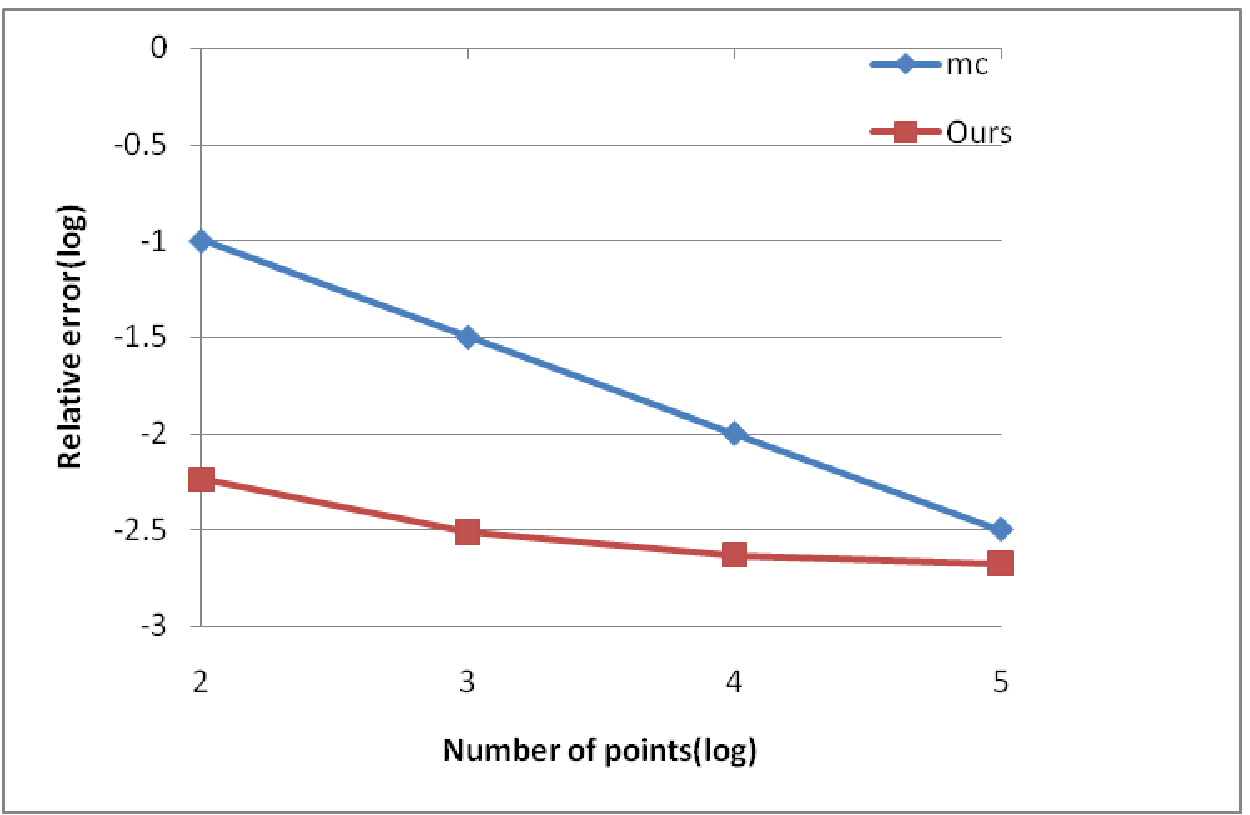}}
\caption{The relative error analysis of calculating the volume of three regular objects for MC and our proposed method.}\label{fig:RegularPoints}
\end{figure*}

\begin{figure*}[ht]
\centering
\subfigure[small case (case 3)]{\includegraphics[height=1in,width=2in]{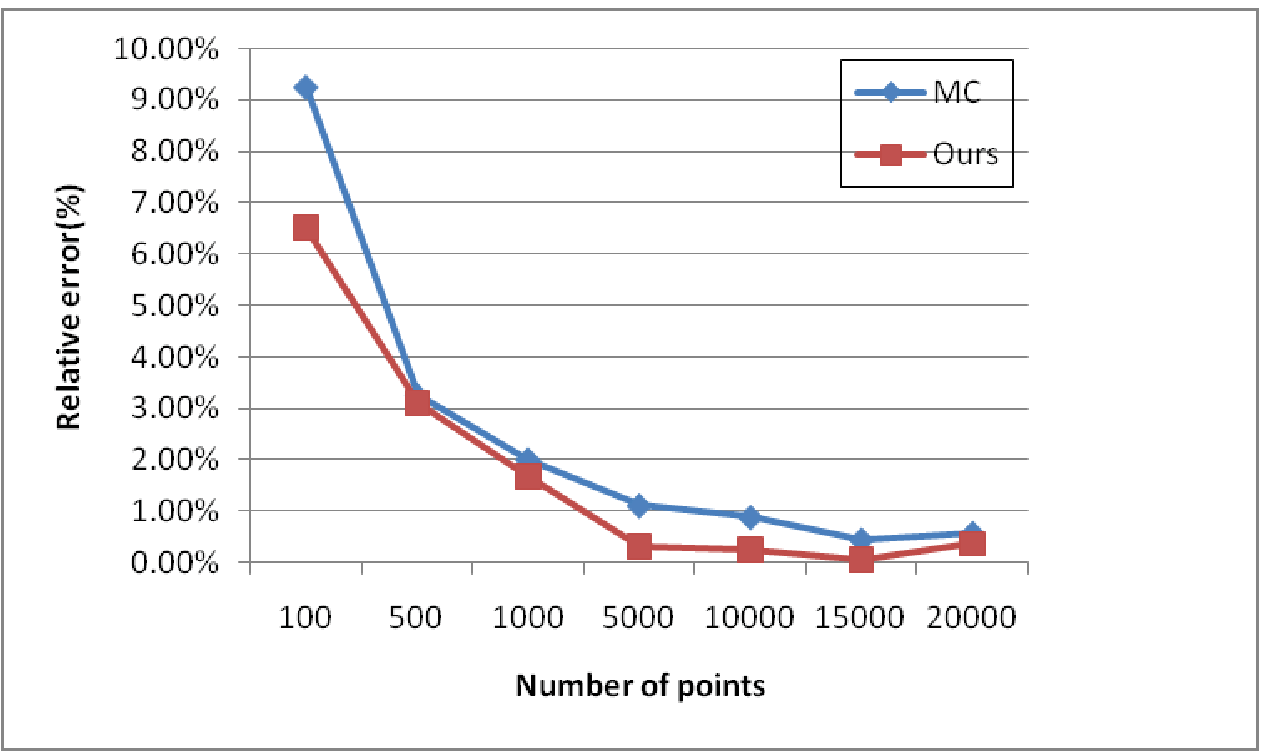}}
\subfigure[medium cases (case 6)]{\includegraphics[height=1in,width=2in]{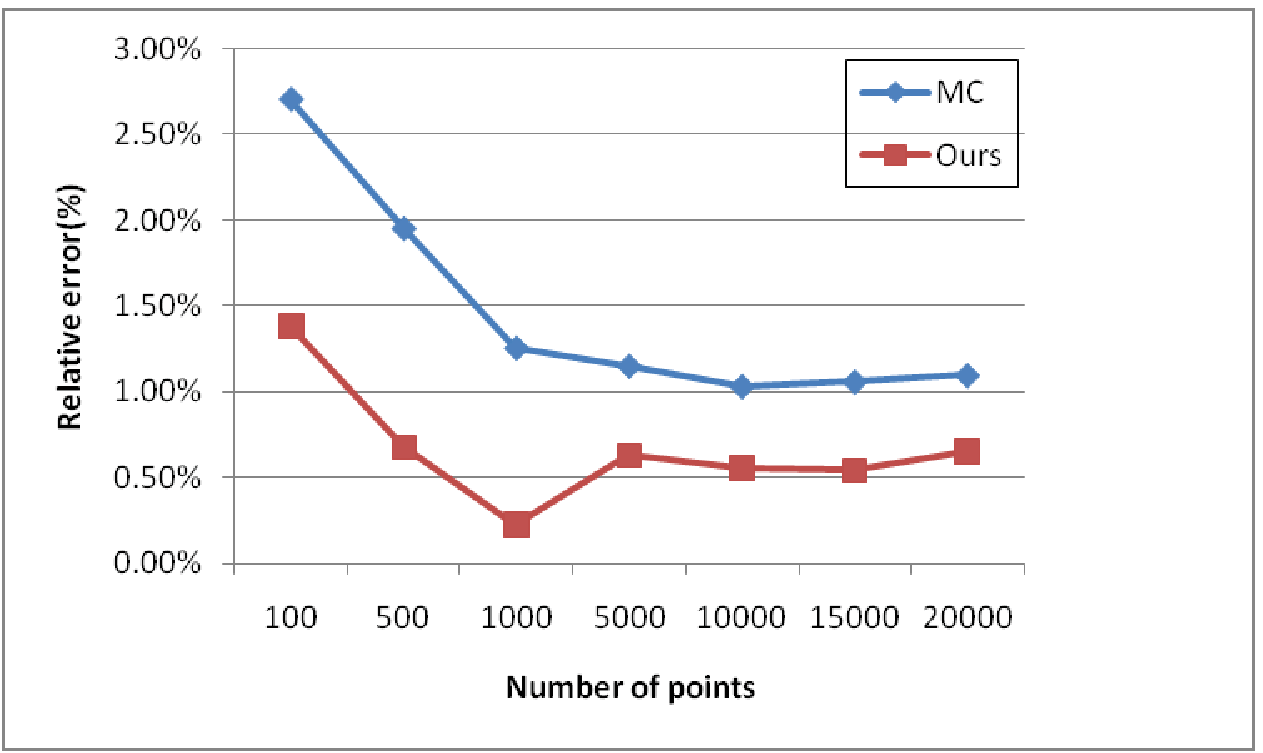}}
\subfigure[large cases (case 9)]{\includegraphics[height=1in,width=2in]{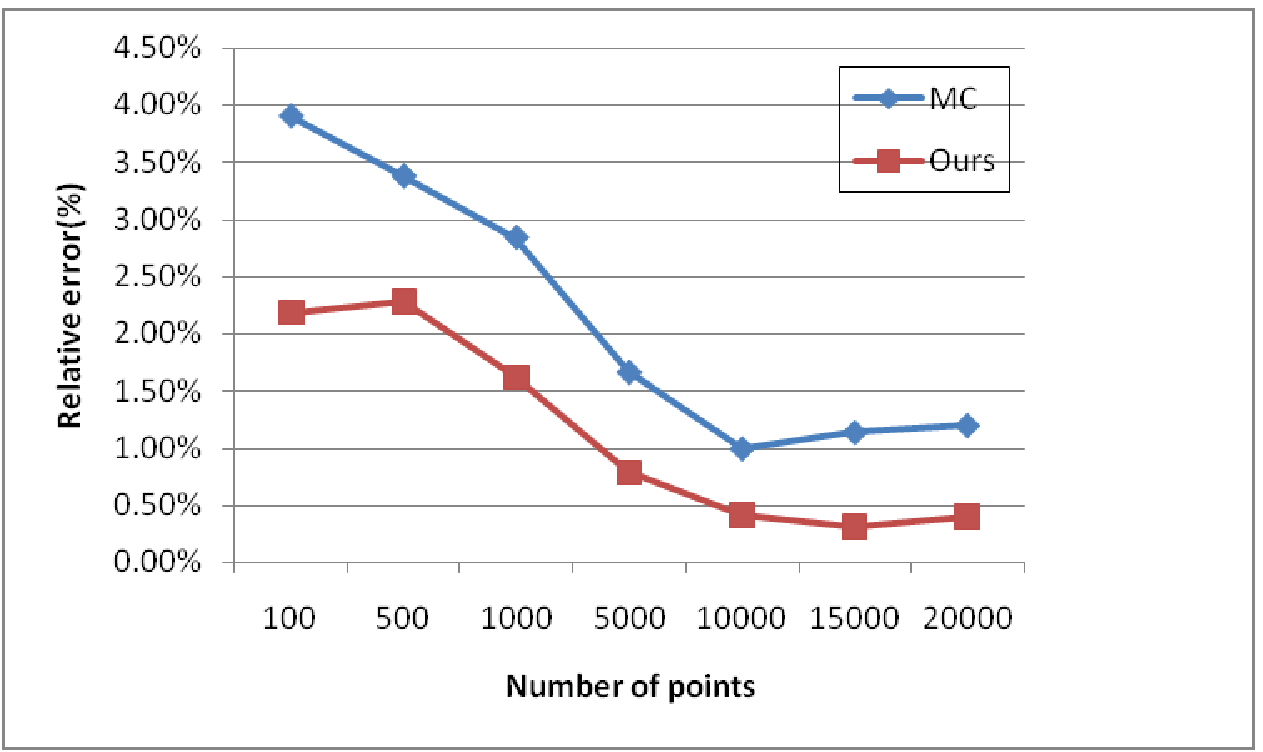}}
\caption{The relative error analysis of calculating the lung lesion volume in three cases for MC and our proposed method.}\label{fig:RE}
\end{figure*}

We also analyze the accuracy of volume calculation when increase the random points for the MC method and our proposed method.
Fig.~\ref{fig:RegularPoints} illustrates the results of obtained of three regular models, the relative error versus the number of points used for volume measurement using the logs to base 10. For each graph, the theoretical error is plotted, which is compared to the actual relative error using our method. In the Fig.~\ref{fig:RE}, we compare the two methods among the three typical cases.

As is shown in the Fig.~\ref{fig:RegularPoints} and Fig.~\ref{fig:RE}. It is illustrated that our proposed method has lower relative errors in all cases compared with MC method. Because for our method, fewer number of points is supposed to achieve a given accuracy.

\begin{table*}[ht]
\centering
\begin{tabular}{l|ccccccccc}
\hline
\hline
Dataset  &True Volume &LSTK (-0.5) &MC (10000) &Ours (10000)\\ \hline
Small(case1) &425.182 &432.334(1.68\%) &417.224(1.87\%) &419.995(\textbf{1.22\%})\\
Small(case2) &1417.62 &1433.16(1.10\%) &1436.13(1.31\%) &1415.85(\textbf{0.13\%})\\
Small(case3) &1533.52 &1551.76(1.19\%) &1505.83(1.81\%) &1537.21(\textbf{0.24\%})\\
Medium(case4) &3444.63 &3489.74(1.31\%) &3518.6(2.15\%) &3415.86(\textbf{0.84\%})\\
Medium(case5) &5234.59 &5305.19(1.35\%) &5165.49(1.32\%) &5186.13(\textbf{0.93\%})\\
Medium(case6) &6820.19 &6898.35(1.15\%) &6836.36(0.24\%) &6782.16(\textbf{0.56\%})\\
Large(case7) &9277.13 &9352.53(0.81\%) &9296.87(0.21\%) &9260.66(\textbf{0.18\%})\\
Large(case8) &9415.15 &9533.36(1.26\%) &9179.48(2.50\%) &9338.65(\textbf{0.81\%})\\
Large(case9) &13634.1 &13758.1(0.91\%) &13635.1(\textbf{0.01\%}) &13577.4(0.42\%)\\
\hline
\end{tabular}
\caption{The results of volume calculation of three methods compared to ground truth volume. The value of each volume is in $mm^3$. MC and our method use 10000 random points. LSTK uses the iso-surface threshold $-0.5$ at its best. The relative error of each volume calculation is shown in brackets after the volume value.}
\label{fig:table}
\end{table*}
Comparing the MC method and ours, the graph above give a distinct analysis. Taking a different look, Table \ref{fig:table} shows the relative efficiency  three regular models and nine cases for the two method. To achieve 1\% accuracy, the number of points approximately needed is shown in the table, and the ratio of MC method and our method is computed.

\begin{table*}[ht]
\centering
\begin{tabular}{l|ccccccccc}
\hline
\hline
Dataset   &MC   &Ours  &Ratio \\ \hline
Cylinder  &350  &100     &3.2      \\
Cuboid    &720   &100    &7.2         \\
Cube      &590  &100     &5.9       \\
Small(case1) &1040 &120 &8.7 \\
Small(case2) &6680 &160 &41.8 \\
Small(case3) &420 &260 &1.6 \\
Medium(case4) &1100 &330 &3.3 \\
Medium(case5) &370 &170 &2.2 \\
Medium(case6) &1350 &110 &12.3 \\
Large(case7) &8610 &110  &78.3 \\
Large(case8) &9110 &10  &911 \\
Large(case9) &230 &110  &2.1 \\
\hline
\end{tabular}
\caption{The number of points approximately needed to achieve an accuracy of 1\%.}
\label{fig:table}
\end{table*}

\section{Discussion and Conclusion}

In this study, the Halton low-discrepancy sequence is applied to calculate the volume for CT lung lesions. Compared with LSTK, the proposed method can calculate volume without reconstructing 3D model of lung lesions and surface triangulating. The proposed method can also generate more uniform random points compared with MC method. The experimental results are demonstrated that our proposed method can achieve more accurate results compared with LSTK and MC method and more stable than MC method.

In our near future study, we will improve the random points selection strategy for volume calculation since the more number of points selected, the more time costs in calculation.

\section*{Acknowledgement}
This work was supported by National Natural Science Foundation of China (Grant No. 61671399, 61327001), Research Fund for the Doctoral Program of Higher Education (20130121120045) and by the Fundamental Research Funds for the Central Universities (20720150110).

% References

\end{document}